\documentclass[twoside,11pt]{article}

\usepackage{jmlr2e}
\usepackage{xcolor}

\jmlrheading{1}{2018}{1-48}{4/00}{10/00}{meila00a}{Przemyslaw Biecek}

\ShortHeadings{DALEX: explainers for complex predictive models}{Przemyslaw Biecek}
\firstpageno{1}

\begin{document}

\title{DALEX: Explainers for Complex Predictive Models in R}

\author{\name Przemys\l{}aw Biecek \email przemyslaw.biecek@gmail.com \\
       \addr Faculty of Mathematics and Information Science\\
       Warsaw University of Technology\\
       75 Koszykowa Street, Warsaw, Poland}

\editor{...}

\maketitle

\begin{abstract}%   <- trailing '%' for backward compatibility of .sty file
Predictive modeling is invaded by elastic, yet complex methods such as neural networks or ensembles (model stacking, boosting or bagging). Such methods are usually described by a large number of parameters or hyper parameters - a price that one needs to pay for elasticity. The very number  of parameters makes models hard to understand. 

This paper describes a consistent collection of explainers for predictive models, a.k.a. black boxes. Each explainer is a technique for exploration of a black box model. Presented approaches are model-agnostic, what means that they extract useful information from any predictive method despite its internal structure. Each explainer is linked with a specific aspect of a model. Some are useful in decomposing predictions, some serve better in understanding performance, while others are useful in understanding importance and conditional responses of a particular variable.

Every explainer presented in this paper works for a single model or for a collection of models. In the latter case, models can be compared against each other. Such comparison helps to find strengths and weaknesses of different approaches and gives additional possibilities for model validation.

Presented explainers are implemented in the DALEX package for R. They are based on a uniform standardized grammar of model exploration which may be easily extended. The current implementation supports the most popular frameworks for classification and regression.
\end{abstract}

\begin{keywords}
machine learning, R, visualization, model interpretability, modeling 
\end{keywords}

\section{Introduction}
In this section we present the importance of model interpretability and state of the art in this domain. Then we show how our methodology helps to better understand complex predictive models a.k.a. black-boxes.

Predictive modeling has a large number of applications in almost every area of human activity, starting from medicine, marketing, logistic, banking and many others. Due to the increasing amount of collected data, models become more sophisticated and  complex. 

It is believed that there is a trade-off between the interpretability and accuracy of a~model (see e.g., \cite{tradeoff_interpretability_accuracy}). It comes from the observation that the most elastic models usually have higher accuracy but in turn they are also more complex. Complexity here means a large number of parameters that affect the final prediction. That number is big enough to make the model ununderstandable for an ordinary human being.

Interpretability may be introduced naturally in the modeling framework, see an example in Figure \ref{frameworkInterpret}.
In many areas the interpretability of a model is very important, see for example \cite{shapley}, \cite{Murad_Tarr}, \cite{Puri_2017}.
The reason behind it that interpretability allows to clash the model structure with the domain knowledge. And this may bring multiple benefits such as:
\begin{itemize}
\item \textbf{Domain validation}. Very flexible models may be over-fitted to the training data and focused on some biases that result from the manner in which the data was collected (sample bias) or some surrogate variables (variable bias). Validation of the model structure helps to identify these biases. In Figure 1 this feature is marked as C.
\item \textbf{Model improvement}. Identification of subsets of observations in which a model has lower performance allows us to correct the model in this subset and leads to further improvements of the model. In Figure 1 this feature was marked as D.
\item \textbf{Trust}. If the model is used to assist people in  activities such as selection of proper therapy,  understanding key factors that drive model predictions is very important.  See more examples in \cite{lime}.
\item \textbf{Hidden debt}. \cite{Dennison_2015} argue that lack of interpretability leads to hidden debt in machine learning models. Despite initial high performance, the real model performance may deteriorate quickly. Model explainers help to control this debt.

\begin{figure}[h!b]
\includegraphics[width=\textwidth]{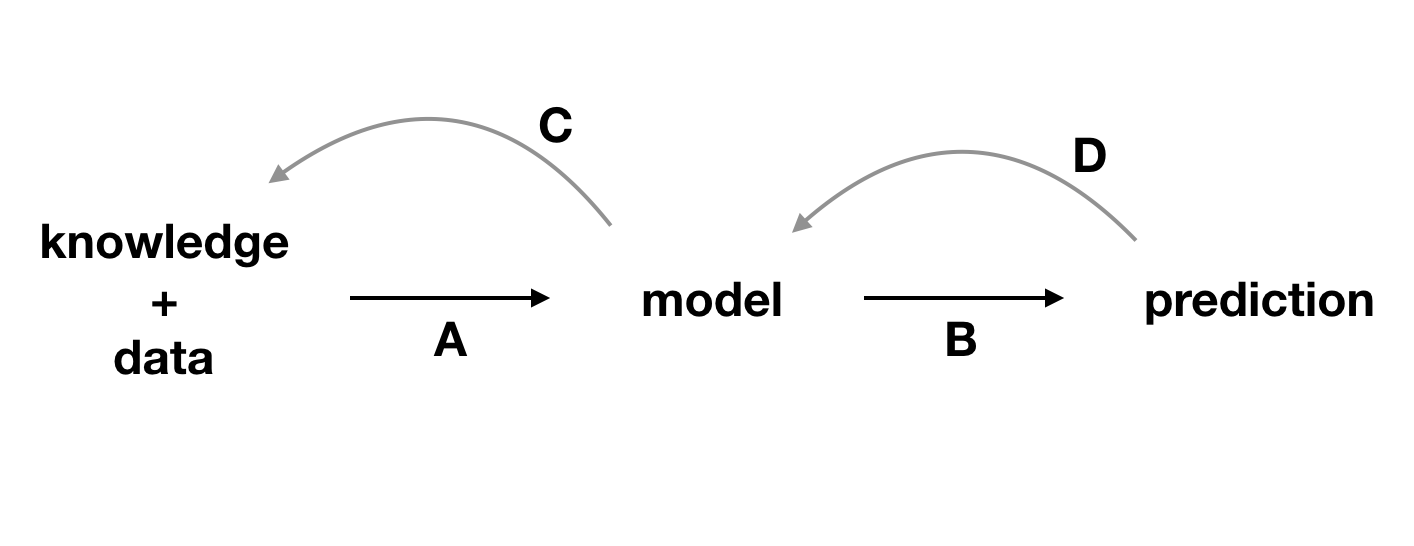}
\caption{\label{frameworkInterpret}Points A and B are from typical workflow of data modeling. A) domain knowledge and data are turned into models. B) models are used to generate predictions, presented methodology extends this framework with new processes
C) model understanding increases our knowledge and, in consequence, it may lead to a~better model, D) undertsanding preditction helps to to correct wrong decisions and, in consequence, it leads to better models.}
\end{figure}

\clearpage

\item \textbf{New insights}. It is hard to increase knowledge about a domain on the basis of black boxes. They may be useful but it does not lead to any new knowledge about a given discipline. Understanding model structure may lead to new interesting discoveries. 

\end{itemize}

In this paper we present a consistent general framework for local and global model interpretability.
This framework covers the most known approaches to model explanation such as Partial Dependence Plots \citep{pdp}, Accumulated Local Effects Plots \citep{ALEPlot}, Merging Path Plots \citep{factorMerger}, Break Down Plots \citep{breakDown}, Permutational Variable Importance Plots \citep{Fisher2018} or Cateris Paribus Plots.

All these explainers are extended in a way that allows us to compare different models against each other on the same scale.
Model comparison is very important since in model building often one gets a collection of competing models. Comparisons of these models and exploration of structures learned by elastic models gives new insights that may be used to construct better features for new models (assisted training with surrogate models). Also lot of effort was put in the graphical side of explainers. Solutions such as Visualizations for Convolutional Networks \citep{Zeiler_Fergus_2014} or Conditional visualization for statistical  models \citep{Connell_Hurley_Domijan_2017} show that well-prepared visualization boost actionability. Also, by purpose, we have not included approaches that do not fit into our grammar of model exploration, such as Individual Conditional Expectations Plot \citep{goldstein2015peeking} and \citep{ALEPlot}. Nevertheless, they are still available, for example in the \texttt{ICEbox} package.

The presented methodology is available as an open source package \texttt{DALEX} for R. The R language \citep{Rcran} is one of the most popular software systems for statistical and machine learning modeling. The current implementation of DALEX supports models generated with the most popular frameworks for classification or regression, such as caret \citep{caret}, mlr \citep{mlr}, Random Forest \citep{randomForest}, Gradient Boosting Machine \citep{gbm} and Generalized Linear Models \citep{glm}. It can be also easily extended to other frameworks and other techniques for model exploration.
The \texttt{DALEX} package is available on GPL-3 license at CRAN\footnote{https://cran.r-project.org/package=DALEX} and at GitHub\footnote{https://github.com/pbiecek/DALEX} along with technical documentation\footnote{https://pbiecek.github.io/DALEX} and extended documentation\footnote{https://pbiecek.github.io/DALEX\_docs}.

Example explainers presented in this paper were recorded with the \texttt{archivist} package \citep{archivist}. Each explainer is an R object, which can be downloaded directly to R console with hooks added to every section. To save space, we present in this paper only graphical representation of explainers. The tabular representation is available through attached hooks. 

\section{Architecture}
\begin{figure}[h!bt]
\centering
\includegraphics[width=0.95\textwidth]{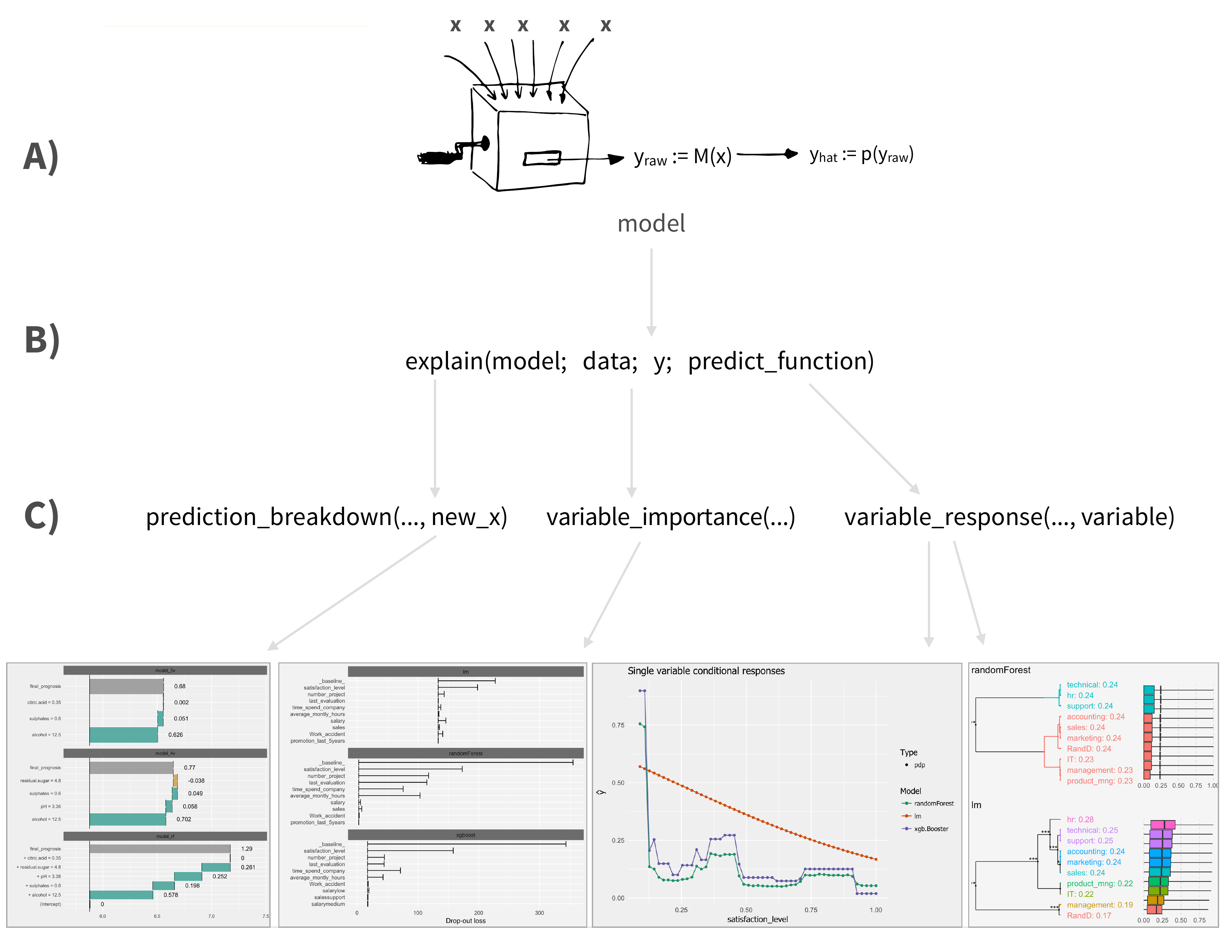}
\caption{\label{fig:architecture}Architecture of the DALEX package is based on simple yet unified grammar. A) Any predictive model with defined input $x$ and output $y_{raw} \in \mathcal R$ may be used. B) Models are first enriched with additional metadata, such as a function that calculates predictions and validation data. The \texttt{explain()} function creates an wrapper over a model that can be used in further processing. C) Various explainers may be applied to a model. 
Each explainer calculates a numerical summaries that can be plotted with generic \texttt{plot()} function.}
\end{figure}

Figure \ref{fig:architecture} presents the general architecture of the DALEX package. The presented methodology is model-agnostic and works for any predictive model that returns a numeric score, such as classification and regression models. 

To achieve a truly model-agnostic solution, explainers cannot be based on model parameters nor model structure. The only assumption here is that we can call the predict function for any selected data points. Such function is wrapped with the model and the validation dataset. Such wrapper serves as a unified interface for a model.

Methods for better understanding of global structure of a model (a.k.a. model explainers) and for better understanding of a local structure of a model (a.k.a. prediction explainers) are implemented in separate functions. We call these function \textit{explainers} since they are design to explain a single feature of a model. As a result, they return numerical summaries in a tabular format. Results from each explainer may be summarized with generic \texttt{plot} function. The \texttt{plot} function will work with any number of models and will overlay all models in a single chart for cross examination.

\clearpage

\section{Model understanding}

In this section we present explainers that increase understanding of a global structure of the model. The primary goal of these explainers is to answer the following questions: How good is the model? Which variables are the most important? How are the variables linked with the model response?

\subsection{Explainers for model performance}

Model performance is often summarized with a single number such as precision, recall, F1, average loss or accuracy. Such approach is handy in model selection. It is easy to construct ranking of models and choose the best one on the basis of a single statistic. 
However, more descriptive statistics are better when it comes to understanding of a model.

The descriptive statistics most often used for classification models is ROC (Receiver Operating Characteristic). It has many various implementations. In R, the most widely used descriptive statistic is the \texttt{ROCR} package \cite{ROCR}. ROC plots have also extensions for regression models. Find an overview of Regression ROC curves in \cite{RROC}.

\begin{figure}[h!tb]
\includegraphics[width=0.49\textwidth]{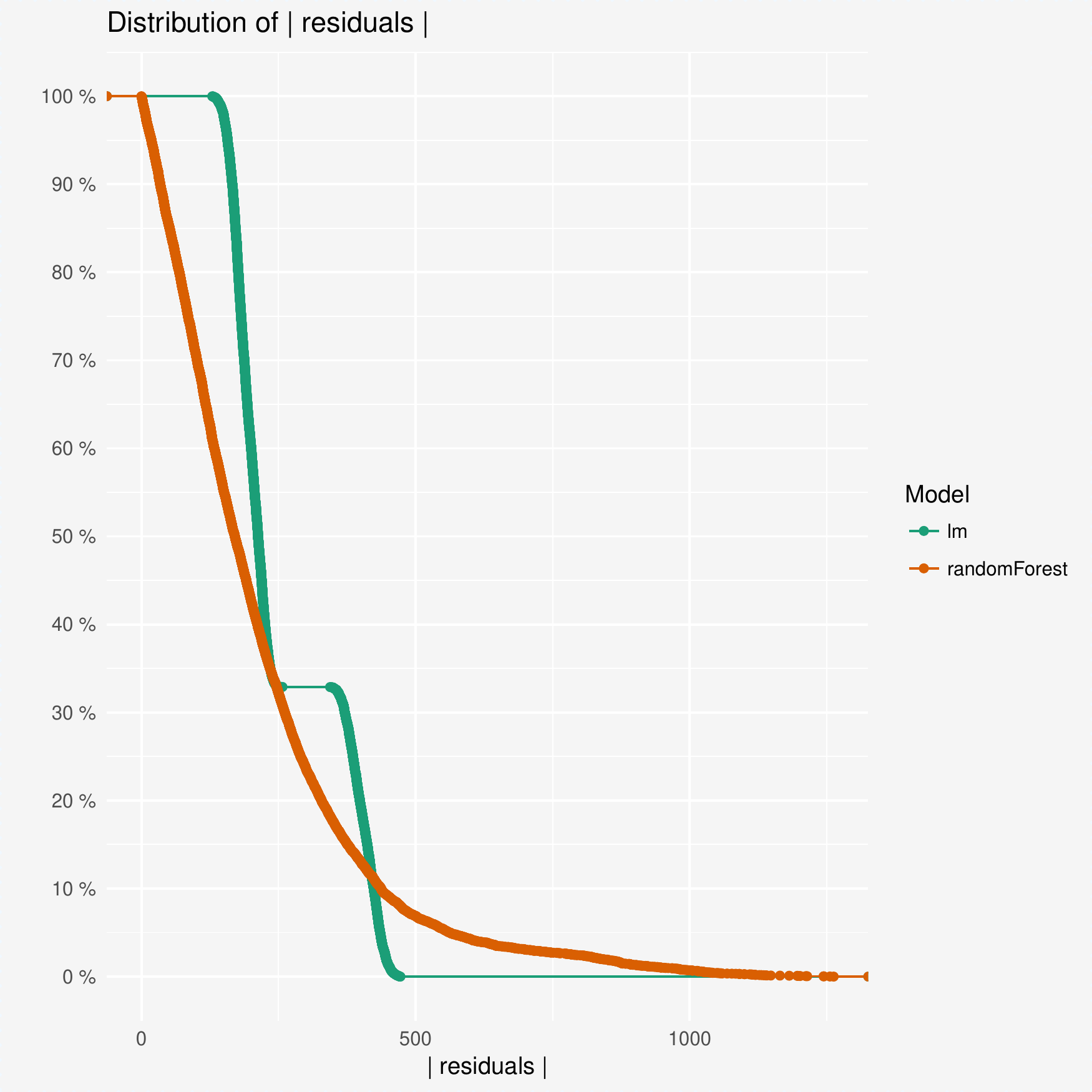}
\includegraphics[width=0.49\textwidth]{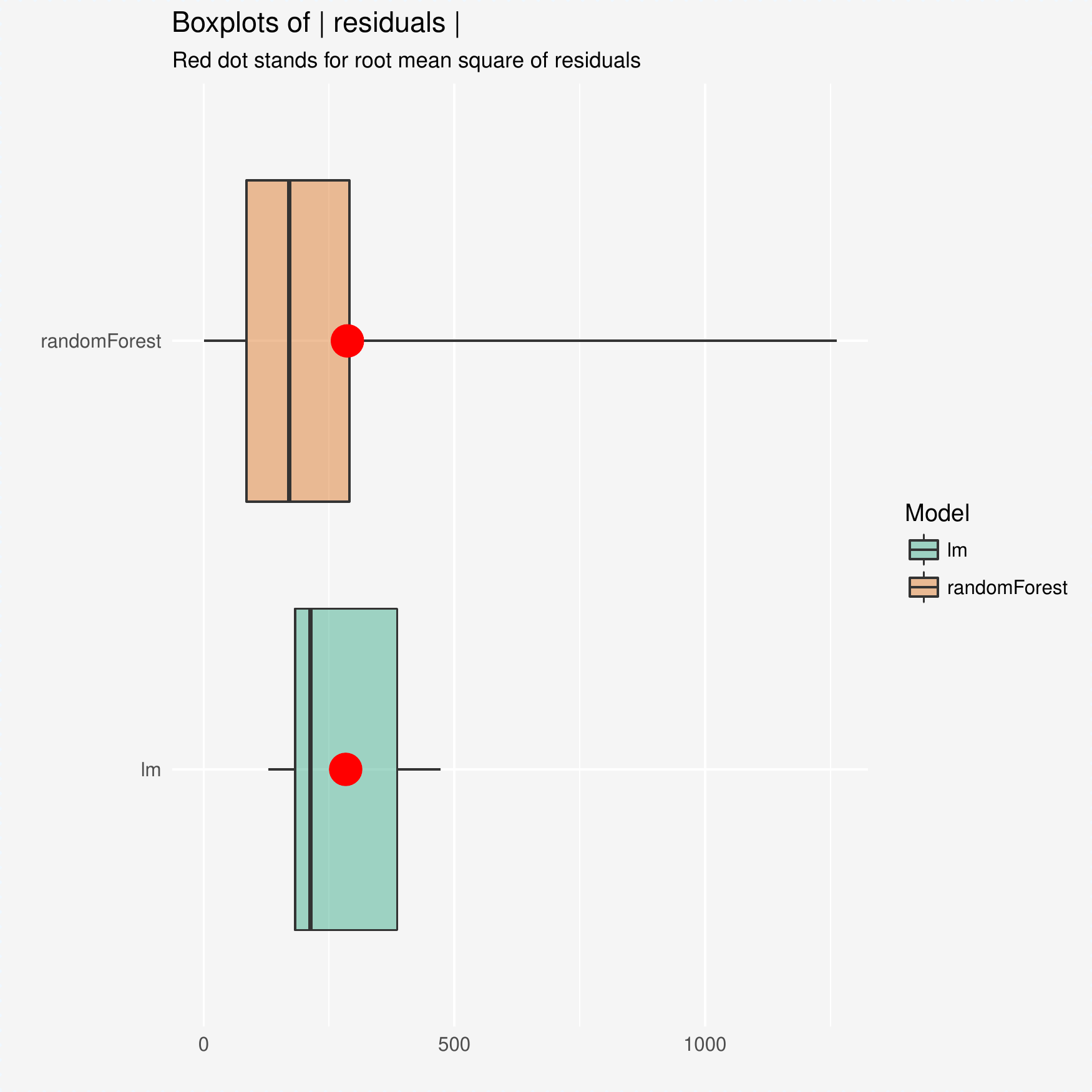}
\caption{\label{modelPerformance1}Both plots compare distributions of residuals for two models. The left plot shows 1 - Empirical Cumulative Distribution Function for absolute values of residuals, while the right plot shows boxplots for absolute values of residuals. Red dots in the right plot stand for root mean square loss.}
\end{figure}

\clearpage

The \texttt{DALEX} package offers a selection of tools for exploration of model residuals.
Figure \ref{modelPerformance1} presents example explainers for model performance\footnote{Access this explainer with  \texttt{archivist::aread('pbiecek/DALEX\_arepo/b4eb1')}} created with \texttt{model\_performance()} function. Here distribution of absolute residuals is compared between two models. The average mean square loss is equal for both models, yet we can see that the random forest model has more small residuals and only a~small fraction of large residuals. 10\% of residuals in random forest model is larger than the largest residual in the linear model.

More diagnostic plots are available through the \texttt{auditor} package \citep{auditor}, which is closely integrated with the \texttt{DALEX}. 

\subsection{Explainers for conditional effect of a single variable}
\label{sec:pdpexpl}

The \texttt{DALEX} package offers a selection of tools for better understanding of a  conditional model's response based on a single variable. Current implementation covers:

\begin{itemize}
    \item Partial Dependence Plot \citep{pdp}, as implemented in the \texttt{pdp} package.
    \item Accumulated Local Effects Plot \citep{ALEPlot} as implemented in \texttt{ALEPlot} package,
    \item Merging Path Plot \citep{factorMerger} as implemented in the \texttt{factorMerger} package.
\end{itemize}
First two methods were designed to deal with continuous variables, while the third one is designed for categorical variables.

\begin{figure}
\includegraphics[width=0.49\textwidth]{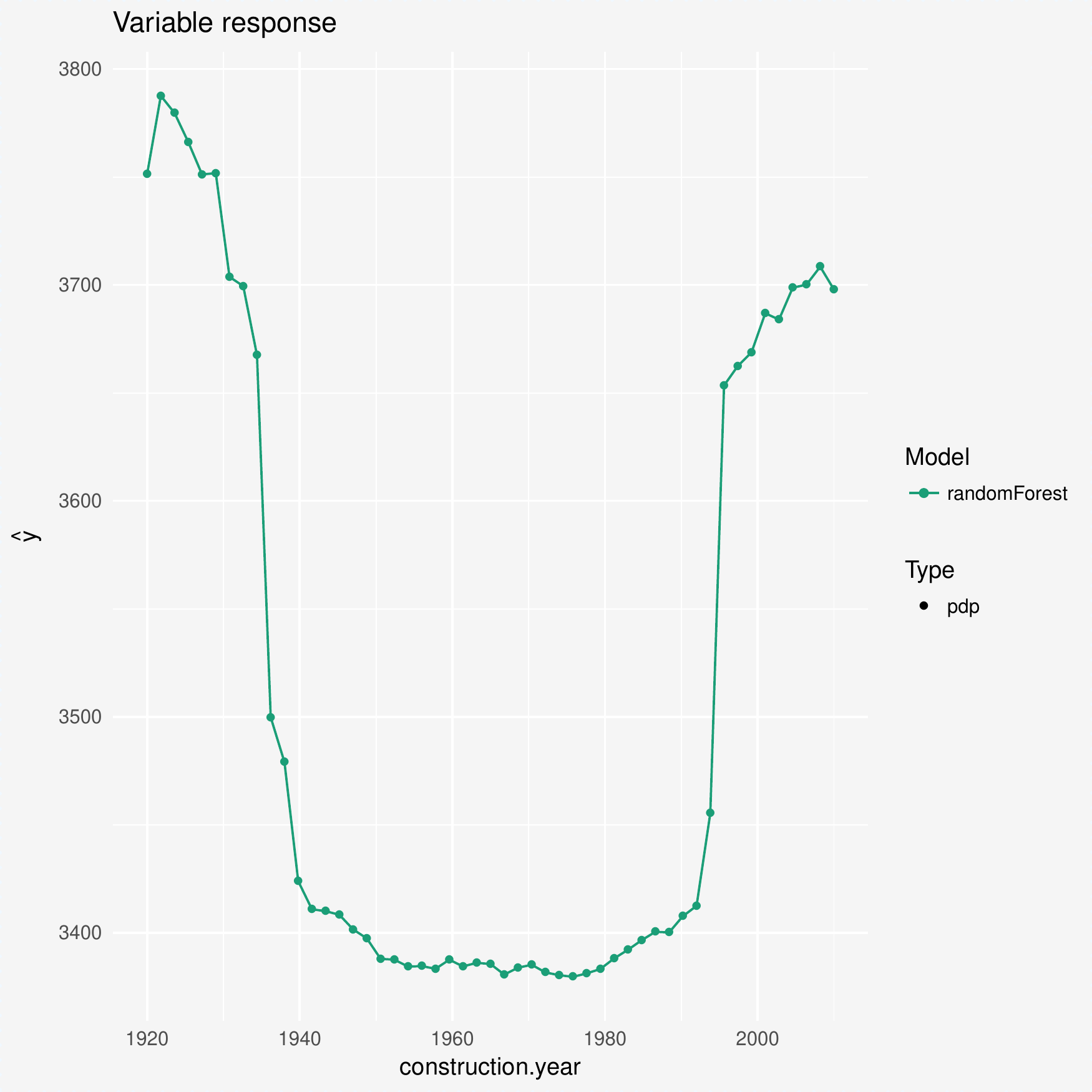}
\includegraphics[width=0.49\textwidth]{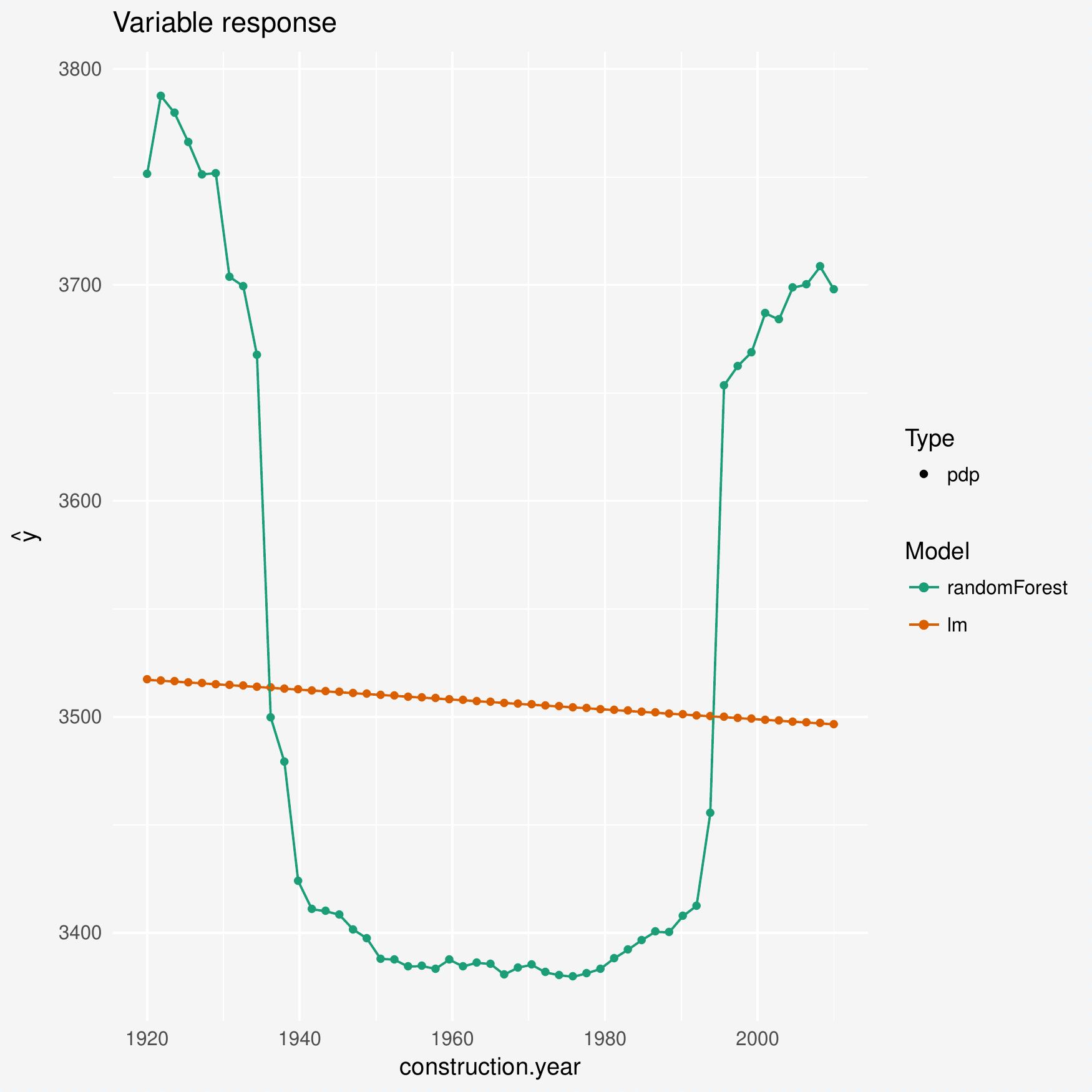}

\includegraphics[width=0.49\textwidth]{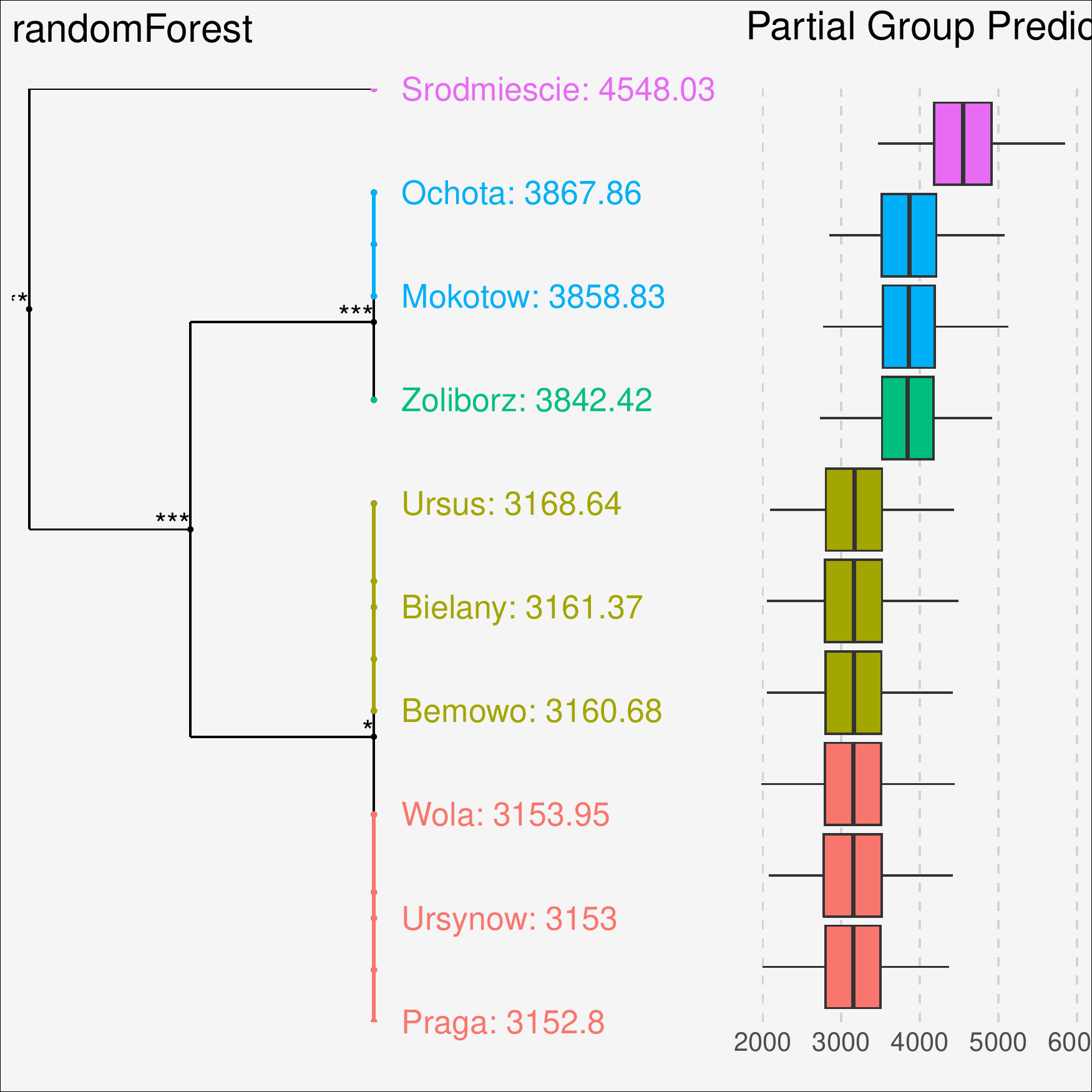}
\includegraphics[width=0.49\textwidth]{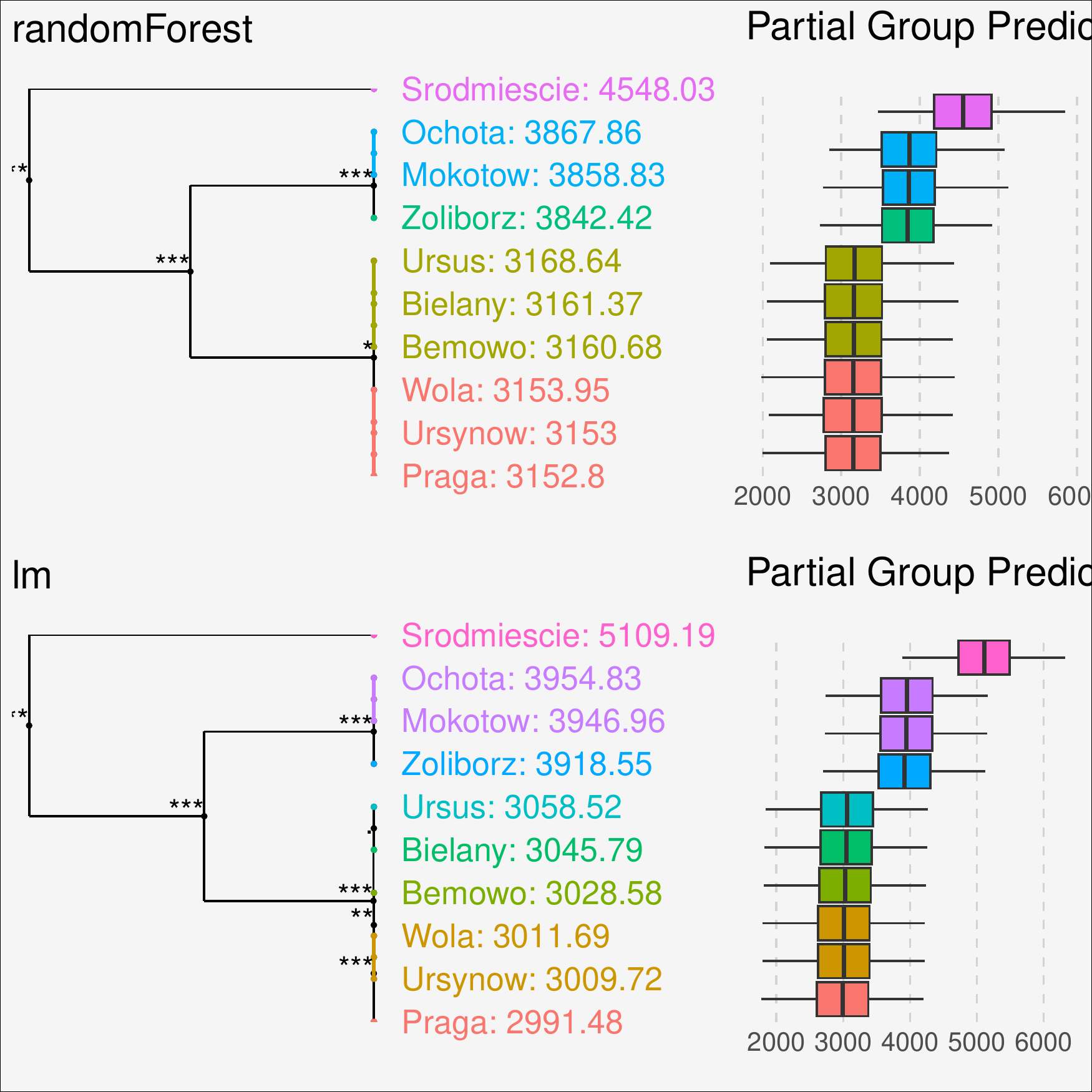}
\caption{\label{single_variable_explainer}Example explainers for variable responses. Two top plots show responses for quantitative variable \texttt{construction year} (Partial Dependency Plots), while bottom plots show responses for factor variable \texttt{district} (Factor Merger Plots). The left plots show explainers for a single model, while the right panels show plots in which two models are being compared. Explainers for quantitative variable show the expected response given a selected value of a variable. Explainers for factor variable present similarity of responses for each possible value.}
\end{figure}

Examples for these explainers\footnote{Access these explainers with  \texttt{archivist::aread('pbiecek/DALEX\_arepo/3b150')} and \texttt{archivist::aread('pbiecek/DALEX\_arepo/6cbf4')}} created with function  \texttt{variable\_response()}  are presented in Figure \ref{single_variable_explainer}.
On the basis of these explainers it is easy to see that the random forest model learns the nonlinear relation between price and construction year. The linear model is unable to handle such relation without some prior feature engineering.
For categorical variable we can see that both models divide the district variable into three groups of values: downtown (largest responses), three districts close to downtown (middle response) and all remaining responses.

\clearpage

\subsection{Explainers for variable importance}

The \texttt{DALEX} package offers a model-agnostic procedure to calculate variable importance. The model-agnostic approach is based on permutational approach introduced initially for random forest \citep{Breiman2001} and then extended for other models by \cite{Fisher2018}.

\begin{figure}[h!bt]
\includegraphics[width=0.49\textwidth]{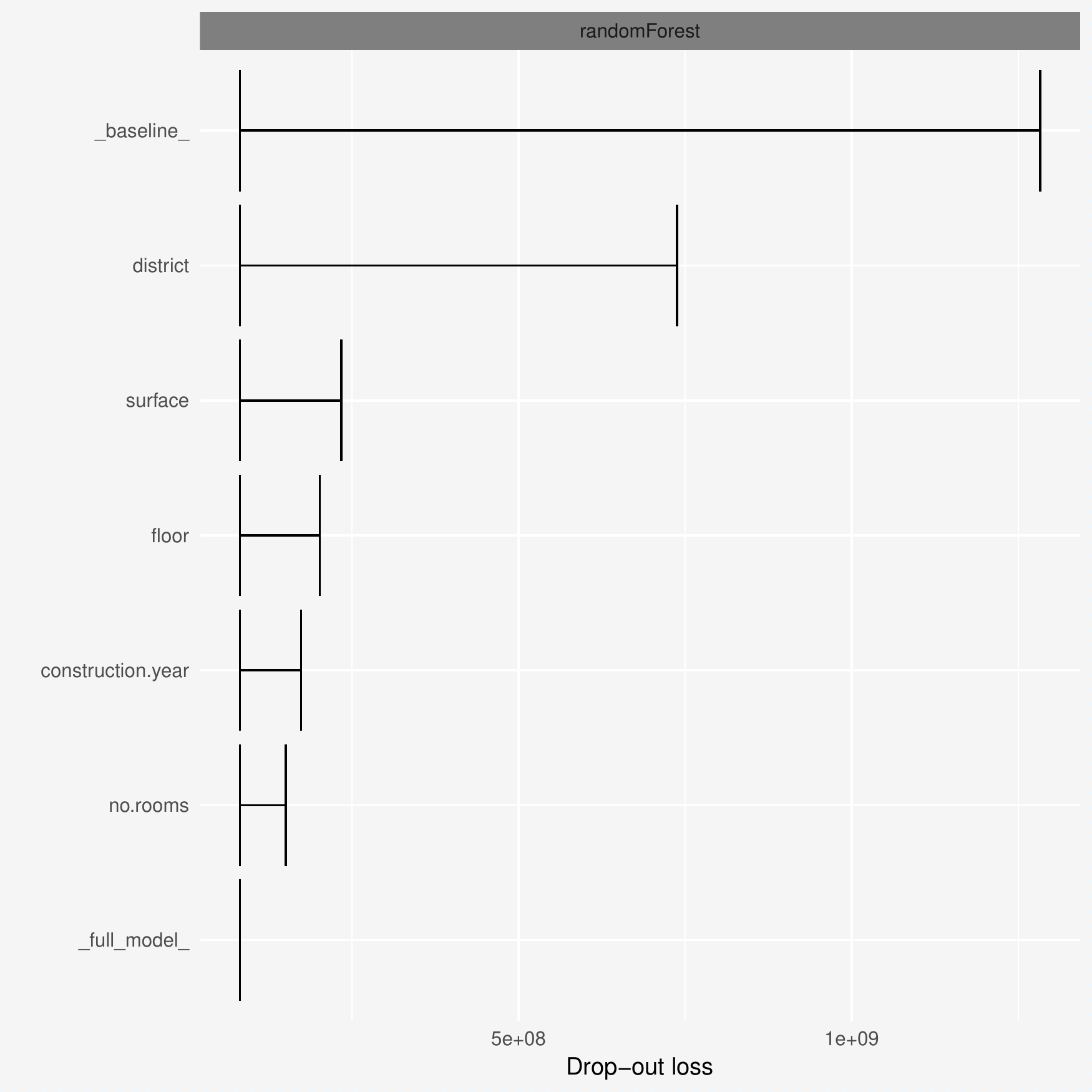}
\includegraphics[width=0.49\textwidth]{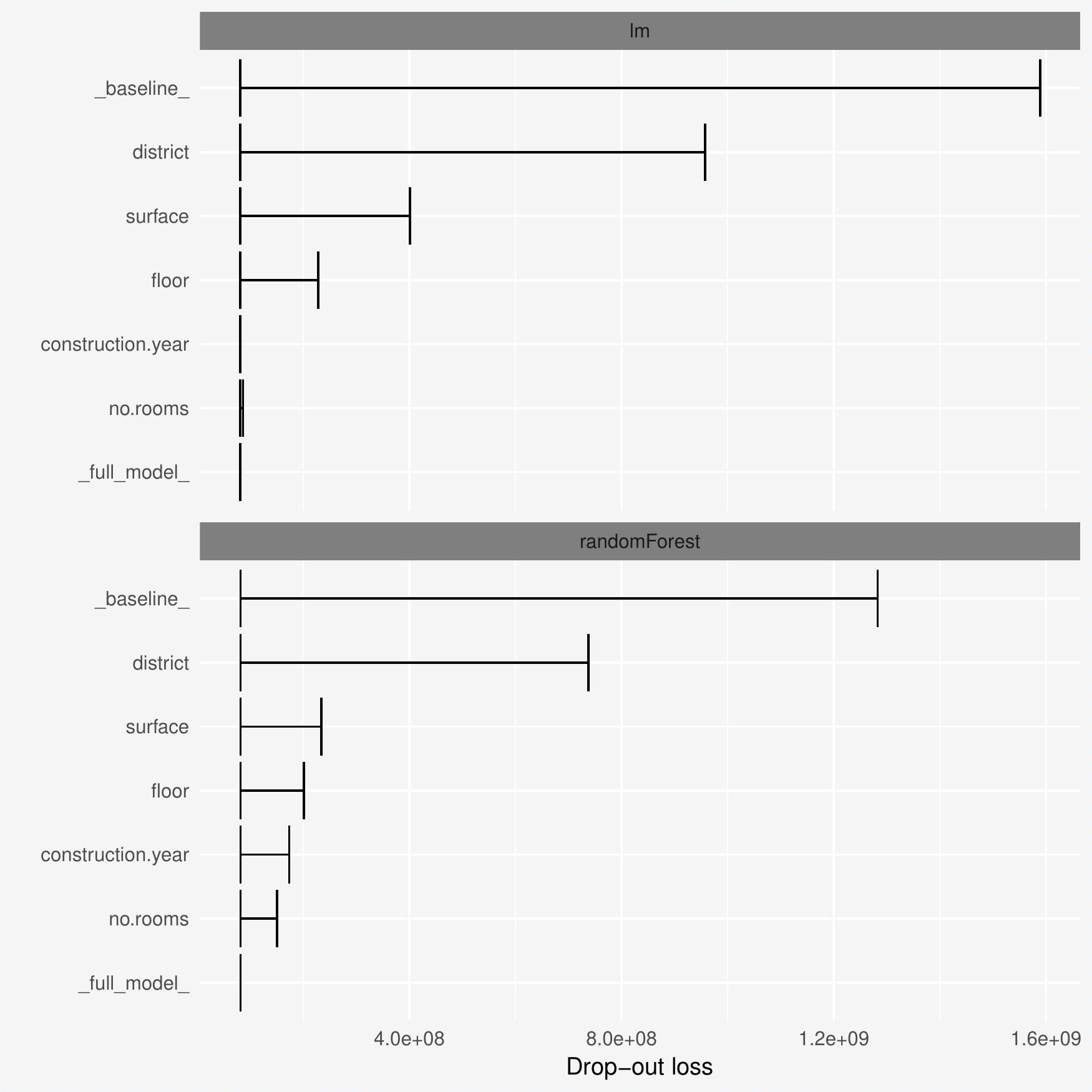}
\caption{\label{variable_importance_explainer}Example explainers for variables importance. Left panel shows explainers for a single model, random forest, while the right one compares two models. Importance of every variable is presented as an interval. One end of this interval is the model performance with regard to validation data, while the second end is the performance with regard to a data set with single variable being shuffled. The longer the interval, the more important the corresponding variable is. }
\end{figure}

An example for these explainers\footnote{Access this explainer with \texttt{archivist::aread('pbiecek/DALEX\_arepo/9378c')}} created with function  \texttt{variable\_importance()} is presented in Figure \ref{variable_importance_explainer}.
The initial performance of both models is similar, and for that reason these intervals are left aligned. For both models the district and surface variables are the most interesting variables. The largest difference between these models is the effect of construction year. For the linear model the length of corresponding interval is almost 0, while for the random forest model is far from 0. This observation is aligned with variables' effects presented in Figure \ref{single_variable_explainer}.

The usual practice in variable importance charts is to present only the length of the interval which is related to loss in the performance metrics after the selected variable is shuffled. Bars on such plots are hitched in 0. 
In the \texttt{DALEX} package we propose to present not only drop in model performance but also the initial model performance. In that way one can compare variables between models with different initial performance.

\clearpage

\section{Prediction understanding}

In this section we present explainers that increase understanding of a prediction for a~single observations. The primary goal of these explainers is to answer the following questions: How stable is the prediction? Which variables influence the prediction? How to attribute effects of particular variables to a single model prediction?

\subsection{Explainers for robustness of predictions}

Ceteris Paribus Plots show how the model response changes as a function of a single variable. These plots recollect similarities to Partial Dependency Plots presented in Section \ref{sec:pdpexpl}; the only difference between them is the fact that Ceteris Paribus Plots are focused on a single observation.

CP Plots have many applications. The derivative is related to local variable importance (as measured in LIME), the profile may be used to verify some constraints related to a~variable (such as monotonic relation) or to asses variable contribution.

\begin{figure}[h!tb]
\includegraphics[width=0.49\textwidth]{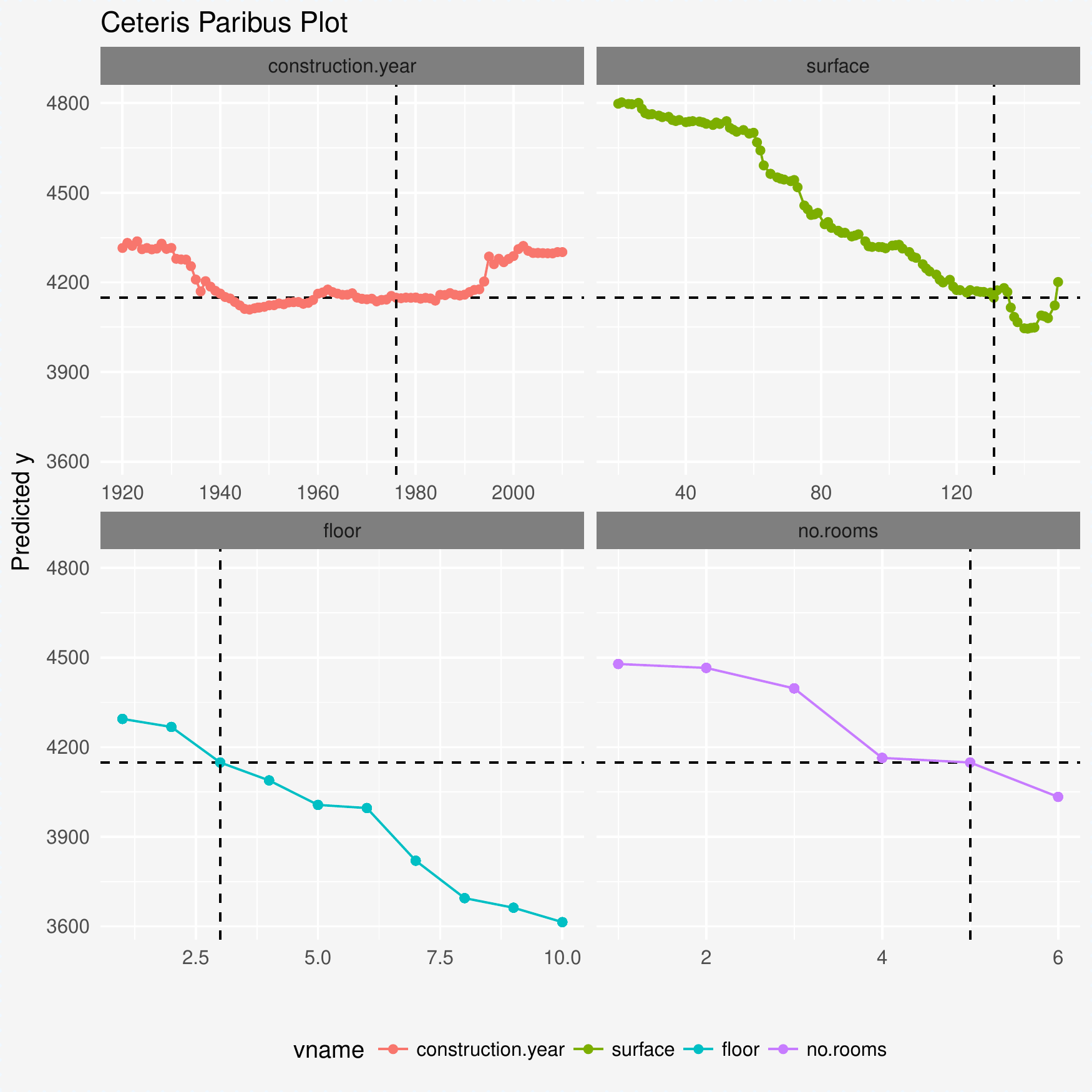}
\includegraphics[width=0.49\textwidth]{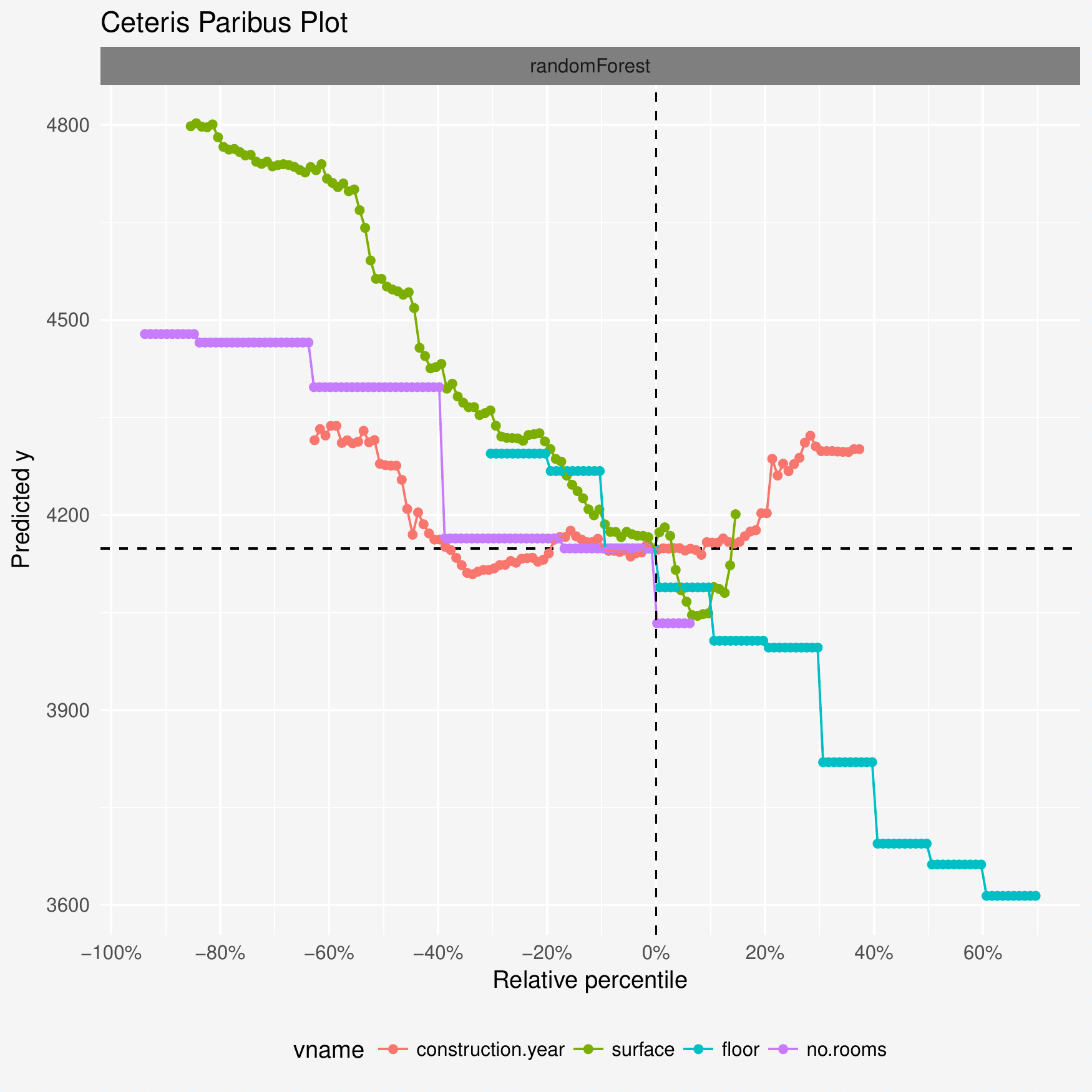}
\caption{\label{ceterisParibus}Ceteris Paribus Plots - explainers for a single observation. The left plot shows how the model response fluctuates for a single observation (predicted y is on OY axis) and if all its unchanged variables remain constant when a single variable is changed (ceteris paribus principle). The right plot shows effects of all variables in the same coordinate system. On OX axis values are normalized through quantile transformation.}
\end{figure}

An example for this explainer\footnote{Access this explainer with \texttt{archivist::aread('pbiecek/DALEX\_arepo/c8989')}} created with  \texttt{ceterisParibus} package\footnote{https://github.com/pbiecek/ceterisParibus} is presented in Figure \ref{ceterisParibus}. We can read from it that the variable \texttt{surface} has the largest effect on the model predictions and and it lowers the model prediction for large apartments. We can also read that small changes in the variable construction year will not affect model predictions.

\clearpage

\subsection{Explainers for variable attribution}

The most known approaches to explanations of a single prediction are LIME method \citep{lime}, working best for local explanations, and Shapley values \citep{Strumbelj2010, Strumbelj2014, shapley}, working best for variable attribution. Break Down Plots are fast approximations of Shapley values. The methodology behind this method and comparison among these three methods is presented in \citep{breakDownlive}.

\begin{figure}[h!tb]
\includegraphics[width=0.49\textwidth]{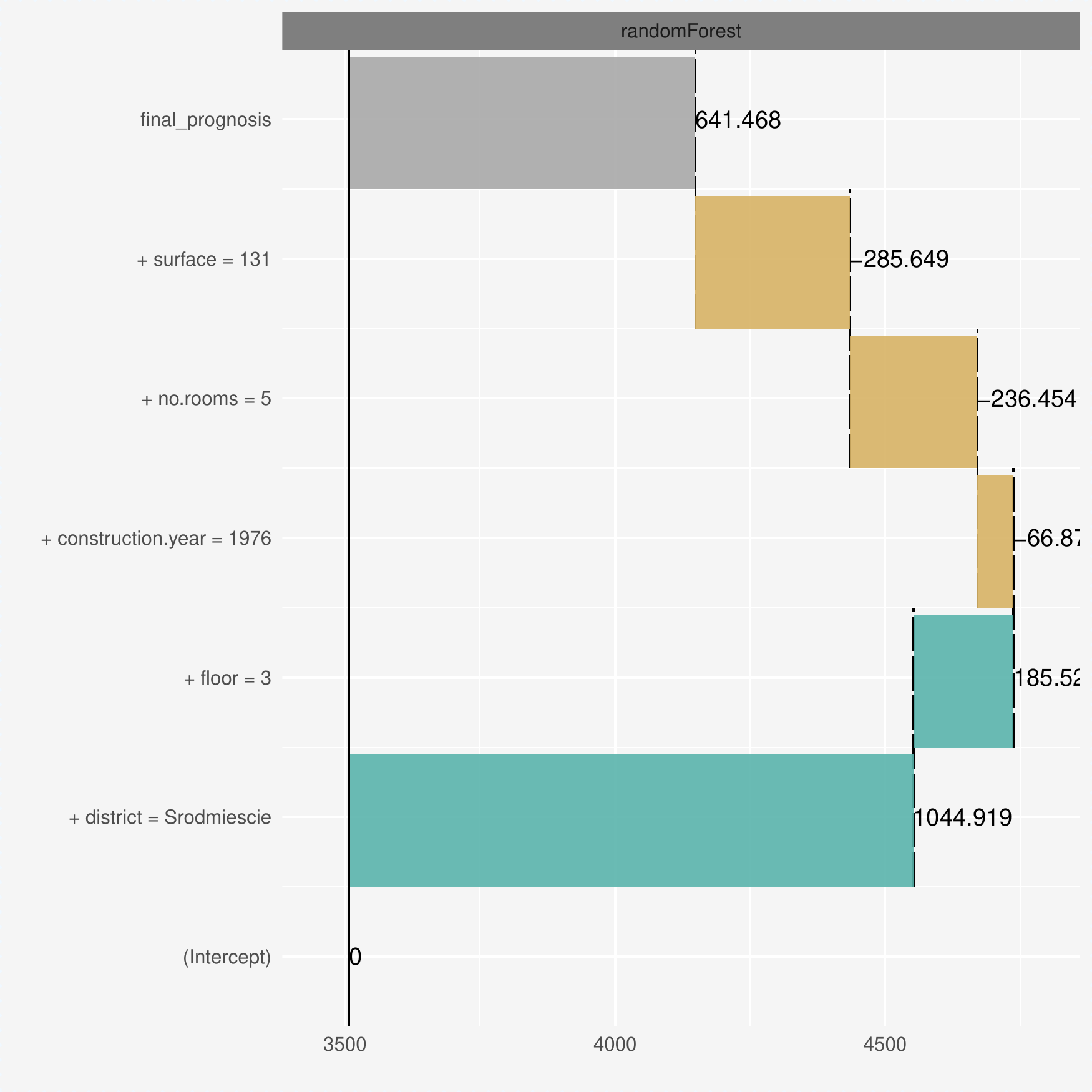}
\includegraphics[width=0.49\textwidth]{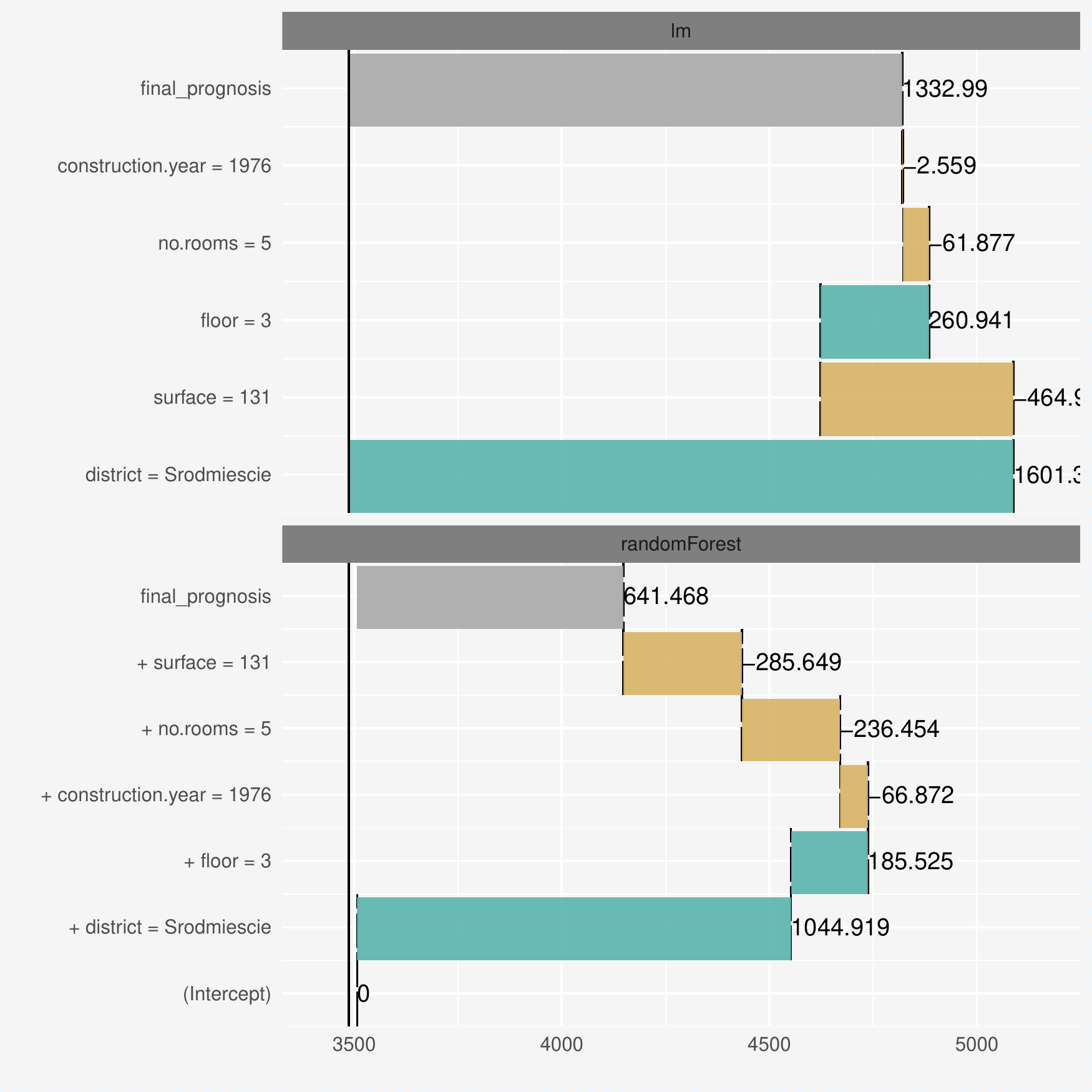}
\caption{\label{breakDown}Break Down Plots - explainers for a single observation that attributes variables to parts of model prediction. The left plot shows how the random forest model's response decomposed onto five variables. The right plot shows decompositions for two models. The gray rectangles show how the single model prediction is different from the population average (reference), the blue rectangles show which variables increase model prediction, while the yellow rectangles are related to variables that lower the model prediction. }
\end{figure}

An example for BDP explainers\footnote{Access this explainer with \texttt{archivist::aread('pbiecek/DALEX\_arepo/72b47')}} created with function \texttt{prediction\_breakdown()}  is presented in Figure \ref{breakDown}.
As one can read from the graph, in both models the largest increase in model prediction is due to variable \texttt{district} = \texttt{Srodmiescie} (downtown). Large surface lowers the prediction in the random forest model, while the variable number of rooms has larger impact in the random forest model than in the linear model.

\section{Summary}

Thinking about data modeling is currently dominated by feature engineering and model training. Kaggle competitions turn the data modeling process into a process that returns a single model with highest accuracy. Tasks of that type may be easily automated. Such thinking about modeling is popular due to lack of tools that can be used for model validation and richer domain verification.

In this article we have introduced consistent methodology and a set of tools for model-agnostic explanations. The presented global explainers for model understanding and local explainers for prediction understanding are based on uniform grammar introduced in Figure \ref{fig:architecture}. Every explainer is constructed in a way that allows for numerical summary, visual summary and comparison of multiple models.

The methodology is developed in a way that is easy to extend with broad technical documentation with rich training materials\footnote{https://pbiecek.github.io/DALEX\_docs}. The code is properly maintained and tested with tools for continuous integration.

\section{Acknowledgments}

The work was partially supported as  RENOIR Project by the European Union Horizon 2020 research and innovation programme under the Marie Sk\l odowska-Curie grant agreement No 691152 (project RENOIR) and by NCN Opus grant 2016/21/B/ST6/02176.

\vskip 0.2in
\bibliography{MLRJ_DALEX}

\end{document}